\documentclass[10pt,twocolumn,letterpaper]{article}

\usepackage[pagenumbers]{wacv} 
\usepackage{colortbl}
\definecolor{darkgreen}{RGB}{0, 154, 85}
\definecolor{darkyellow}{RGB}{255, 204, 0}
\usepackage{multirow}

\definecolor{wacvblue}{rgb}{0.21,0.49,0.74}
\usepackage[pagebackref,breaklinks,colorlinks,allcolors=wacvblue]{hyperref}

\def\wacvPaperID{1031} 
\def\confName{WACV}
\def\confYear{2026}

\title{RelMap: Enhancing Online Map Construction with Class-Aware Spatial Relation and Semantic Priors }

\author{
Tianhui Cai, 
Yun Zhang,
Zewei Zhou,
Zhiyu Huang\thanks{Corresponding author. \texttt{zhiyuh@ucla.edu}}, \
Jiaqi Ma \\ [0.1cm] 
University of California, Los Angeles
}

\begin{document}
\maketitle
\begin{abstract}
Online high-definition (HD) map construction is crucial for scaling autonomous driving systems. While Transformer-based methods have become prevalent in online HD map construction, most existing approaches overlook the inherent spatial dependencies and semantic relationships among map elements, which constrains their accuracy and generalization capabilities.
To address this, we propose \textbf{RelMap}, an end-to-end framework that explicitly models both spatial relations and semantic priors to enhance online HD map construction. Specifically, we introduce a \textbf{Class-aware Spatial Relation Prior}, which explicitly encodes relative positional dependencies between map elements using a learnable class-aware relation encoder. Additionally, we design a Mixture-of-Experts-based \textbf{Semantic Prior}, which routes features to class-specific experts based on predicted class probabilities, refining instance feature decoding.
\textbf{RelMap} is compatible with both single-frame and temporal perception backbones, achieving state-of-the-art performance on both the nuScenes and Argoverse 2 datasets.
\end{abstract}


\section{Introduction}
\label{sec:intro}


\begin{figure}[t!]  
    \centering
    \includegraphics[width=\columnwidth]{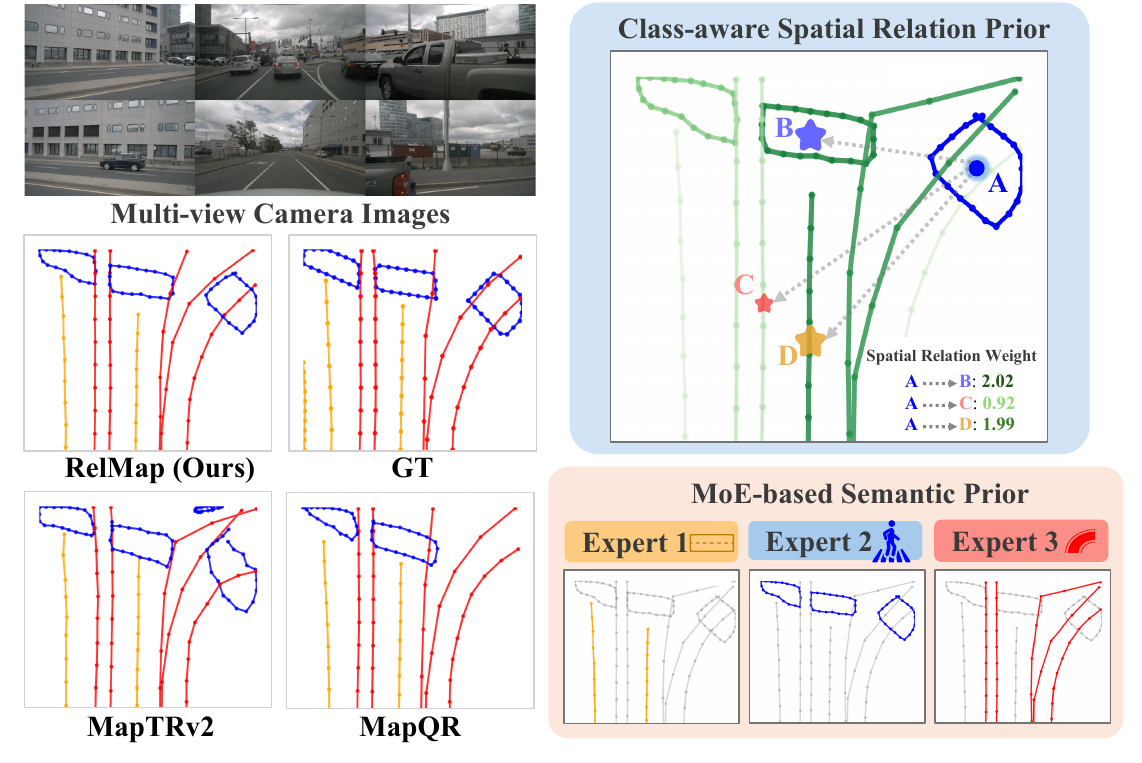}
    \caption{Two major improvements proposed in \textbf{RelMap} for online vectorized map construction. The \textbf{Class-aware Spatial Relation Prior} explicitly encodes spatial dependencies between map elements and decodes important weights, while the \textbf{MoE-based Semantic Prior} routes features to class-specific experts based on predicted class probabilities for refined decoding.}
    \label{fig:hero}
    \vspace{-0.5cm}
\end{figure}

High-definition (HD) maps encode rich geometric and semantic information about road infrastructure and play a pivotal role in autonomous driving systems, enabling localization \cite{gao2024cooperative}, prediction \cite{shi2022motion,shi2024mtr++,sun2023qcnet,zhou2024v2xpnp}, and planning \cite{hu2023planning, Huang_2023_gameformer}. However, the maintenance and frequent updating of HD maps impose substantial costs. Furthermore, the reliance on pre-constructed HD maps significantly limits the scalability and adaptability of autonomous vehicles, particularly in dynamically changing environments. To mitigate these limitations, recent works have explored online HD map construction using onboard sensory data \cite{liu2023vectormapnet,liu2024mgmap, li2022hdmapnet, liao2022maptr,ding2023pivotnet, yuan2024streammapnet, shi2024globalmapnet}. These methods generate HD maps in real-time using perception inputs, thus obviating the need for pre-annotated maps and enhancing scalability in real-world scenarios.

Most existing methods adopt a bird’s-eye-view (BEV) perception backbone \cite{zhou2022cross,chen2022gkt, philion2020lift, wu2024blos} to transform 2D image-space features into BEV features, followed by DETR-like \cite{carion2020end} models to predict vectorized map instances \cite{liao2022maptr, liu2024mapqr, homap, zhou2024himap, yang2024mgmapnet}. Despite their success, these approaches typically treat all map elements uniformly, overlooking the structured priors and spatial-semantic regularities inherent in map topology. While this flexibility may benefit object detection, it is suboptimal for map construction, where elements such as lanes, road boundaries, and crosswalks show consistent geometric patterns and contextual dependencies.

To address this limitation, we propose \textbf{RelMap} that explicitly incorporates spatial and semantic priors into Transformer-based decoders. Our design is motivated by two key insights. First, map elements exhibit structured spatial relationships; for instance, lane dividers often run parallel to road boundaries, and pedestrian crossings are usually positioned near intersections. To leverage such spatial regularities, we introduce a learnable \textbf{Class-aware Spatial Relation Prior} that models inter-instance geometric dependencies. Specifically, we develop a \textbf{Position Relation Encoder} inspired by relation encoding in vision Transformers \cite{li2023deep, liu2023goggle,hu2018relation,hou2024relation}, which computes relative positions between instances and propagates this information across Transformer layers. 
In addition to spatial dependencies, map elements also exhibit class-specific semantic characteristics. For example, crosswalks typically appear as rectangular patterns, while lane dividers are elongated and linear. To capture such class-conditioned variations, we propose a \textbf{Class Relation Modeling Prior}, implemented via a learnable relation matrix that encodes semantic affinities among map classes. This matrix modulates the spatial relation embeddings with class-specific cues, enabling the model to adaptively focus on semantically relevant spatial contexts.
Second, to further enhance class-aware semantic decoding, we introduce a \textbf{Class-conditioned Mixture-of-Experts (MoE)-based Semantic Prior}. Rather than decoding all map instances within a shared feature space, our approach dynamically routes each instance to a specialized expert based on its predicted class probability. Each expert learns to model the distinct appearance and structural characteristics of a specific class, thereby promoting more accurate and semantically consistent decoding. This MoE mechanism, inspired by prior works in dynamic routing \cite{shazeer2017moe}, enables scalable learning of class-specific decoding functions while preserving model efficiency.

To validate the effectiveness of the proposed priors, we build two variants of RelMap: a single-frame model (\textbf{RelMap-SF}) and a temporal model with historical memory integration (\textbf{RelMap-TF}). These two variants demonstrate the applicability of our priors across different temporal settings and architectures, consistently improving vectorized map prediction quality.
The key improvements of our online vectorized HD map construction model are illustrated in \cref{fig:hero}. The main contributions of this paper are summarized as follows:
\begin{enumerate}
\item We introduce two learnable priors in Transformer decoders for map construction: a \textbf{Class-aware Spatial Relation Prior} for explicit modeling of geometric dependencies, and a \textbf{Class-conditioned MoE Semantic Prior} for class-specific decoding.
\item We demonstrate the generality of our approach by integrating the proposed priors into both single-frame and temporally-aware mapping models.
\item Our approach achieves state-of-the-art performance across both single-frame and temporal settings in real-world online HD map construction benchmarks. 

\end{enumerate}
\section{Related Work}

\noindent \textbf{Vectorized Map Construction.}
HD map construction aims to generate structured maps from sensor data. Early models like HDMapNet \cite{li2022hdmapnet} rely on BEV segmentation and post-processing, while VectorMapNet \cite{liu2023vectormapnet} introduces a two-stage, auto-regressive pipeline. MapTR \cite{liao2022maptr} and MapTRv2 \cite{liao2024maptrv2} enable end-to-end vectorized prediction via hierarchical queries, inspiring a series of follow-up works \cite{bemapnet, zhang2024gemap, yang2024mgmapnet, homap, wang2024priormapnet, yuan2024streammapnet}. Recent works such as StreamMapNet \cite{yuan2024streammapnet} and MapTracker \cite{chen2024maptracker} further enhance performance by leveraging historical context through memory aggregation, temporal query modeling, or feature propagation \cite{yuan2024streammapnet, wang2024stream, chen2024maptracker, yang2025histrackmap, peng2025prevpredmap, 10943317, peng2025uni, kim2024unveilinghiddenonlinevectorized}.

Several prior-based methods have been proposed to improve HD map construction. Some leverage external data like SD maps \cite{jiang2024sdhdprior, luo2024smurfsd, zhang2024enhancing, gao2024complementing, peng2025uni}, while others learn priors to guide training \cite{liu2024mgmap, wang2024priormapnet, zhang2024hybrimap}. For example, MGMap \cite{liu2024mgmap} introduces instance- and position-guided masks to refine feature localization, and PriorMapNet \cite{wang2024priormapnet} integrates structure priors from clustered map elements into query initialization. However, these works neglect spatial relationship priors across instances. We address this by introducing a learnable, class-aware spatial relation prior that promotes more structured and context-aware predictions.


\noindent \textbf{Relation Modeling.}
Relation modeling has demonstrated effectiveness across various computer vision tasks, including image recognition \cite{chen2019multi}, object detection \cite{jiang2018hybrid}, and generative modeling \cite{li2023deep, liu2023goggle}. Category-level modeling captures class co-occurrence patterns \cite{chen2019multi, hao2023relation}, while instance-level methods focus on spatial or semantic relations between objects \cite{li2020gar, lin2021core, hou2024relation, hu2018relation}.
Despite the effectiveness of relation modeling in vision tasks, its application in HD map construction remains underexplored \cite{shin2025instagram, xu2024insmapper, luo2025reltopo, hu2025fastmap}, particularly explicit spatial relation modeling. GeMap \cite{zhang2024gemap} separates attention into shape attention for intra-instance refinement and relation attention for inter-instance interactions, but lacks explicit relation modeling conditioned on classes. We introduce an instance-level spatial relation modeling approach that incorporates class-aware adjustments, enabling the model to capture spatial dependencies more effectively across different map categories.

\noindent \textbf{Mixture-of-Experts.}
Mixture-of-Experts (MoE) \cite{shazeer2017moe} enhances model capacity expressiveness by dynamically routing inputs to specialized sub-networks (“experts”). Instead of processing all inputs through a shared network, MoE assigns each input to selected experts, typically FFNs, via a gating function. Sparse gating selects only a subset of experts per input, often using a top-K strategy \cite{fedus2022switch, zoph2022st, lewis2021base, dai2022stablemoe, dai2024deepseekmoe}, whereas dense gating activates all experts with different weights \cite{pan2024dense, wu2024mixture, dou2023loramoe}.

MapExpert \cite{zhang2024mapexpert} employs a sparse MoE \cite{shazeer2017moe} with a top-K gating mechanism, where each input is assigned to a subset of experts with the highest gating scores. Like traditional MoE approaches, it relies on a router network for expert selection and an auxiliary expert balance loss to prevent imbalanced expert utilization. In contrast, our method eliminates the need for a separate router and expert balance loss by directly leveraging the classification predictions from the previous decoder layer. This simplifies the training process, reduces additional parameters, and achieves better performance with fewer experts.

\begin{figure*}[tbp]  
    \centering
    \includegraphics[width=\textwidth]{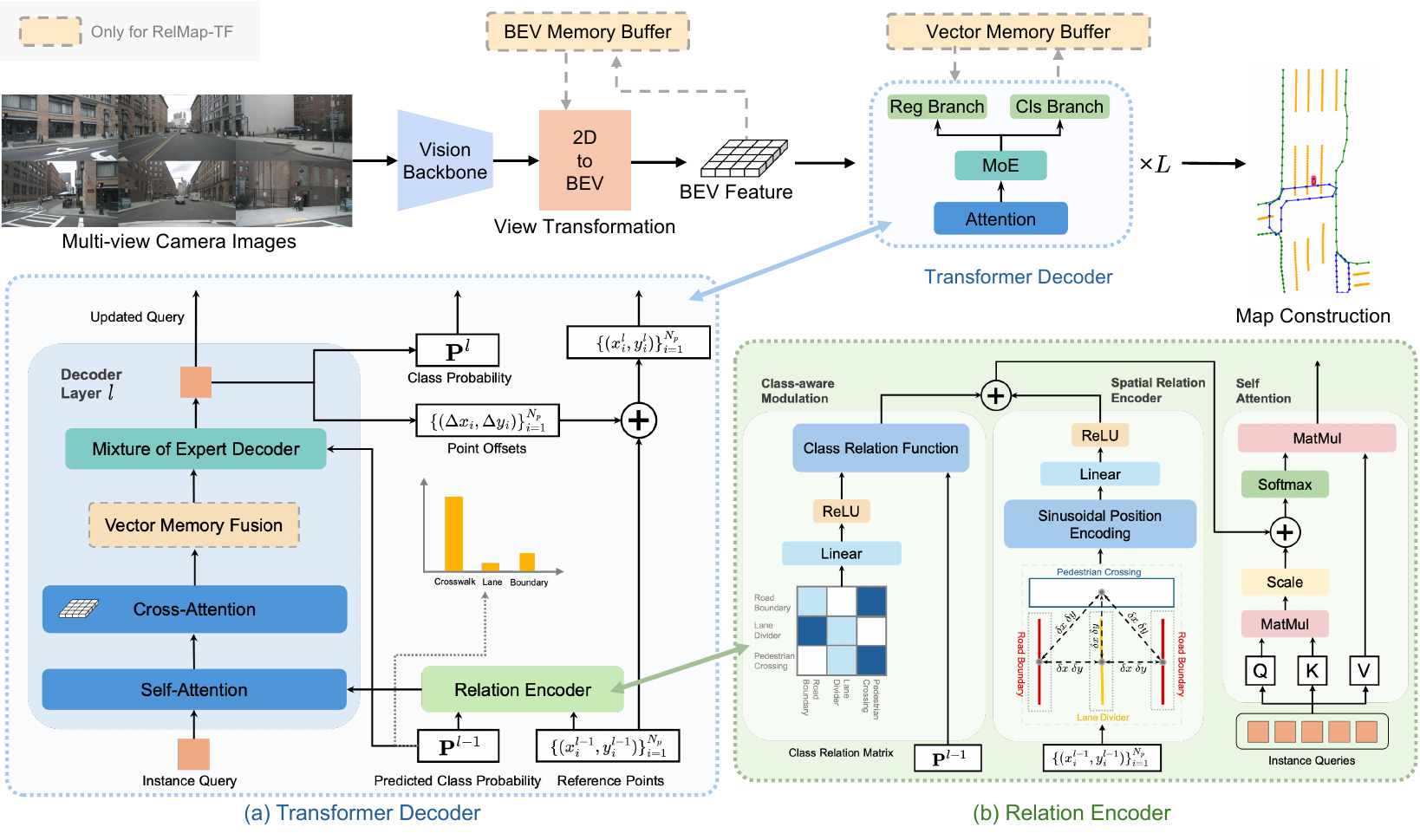}  
    \caption{Overview of \textbf{RelMap}, our proposed online vectorized map construction framework. Given multi-view camera images, the model first extracts image features using a vision backbone and transforms them into the BEV space. Subsequently, an enhanced Transformer decoder predicts vectorized map elements. In the temporal variant, RelMap-TF, memory buffers and a fusion module integrate historical context. To improve spatial and semantic reasoning, we refine the decoder's self-attention mechanism using a class-aware relation encoder and introduce a MoE decoder, where class probabilities guide the routing of map instances to specialized expert networks. }
    \label{fig:framework}
    \vspace{-0.4cm}
\end{figure*}


\section{Method}

In vectorized HD map construction, we construct map instances from surrounding perspective-view (PV) images. Each map instance is represented as a vector with an ordered set of 2D points $\{(x_i, y_i)\}_{i=1}^{N_p}$, where $N_p$ is the number of points per instance. 

\subsection{Main Framework}

To enhance spatial and semantic reasoning in vectorized map construction, as illustrated in \cref{fig:framework}, we introduce two learnable priors: (1) a Learnable Class-aware Spatial Relation Prior, which enhances spatial dependency modeling between map instances, and (2) a MoE-based Semantic Prior, which enables class-specific feature refinement.

We instantiate these priors in two model variants: \textbf{RelMap-SF}, a single-frame map construction model built on MapQR \cite{liu2024mapqr}, and \textbf{RelMap-TF}, a temporal map construction model built on MapTracker \cite{chen2024maptracker}. Although these models differ in specific architectural components, they share a common foundation inherited from the MapTR series \cite{liao2022maptr, liao2024maptrv2}, including: (1) a ResNet-50 \cite{he2016resnet} vision backbone for multi-view image feature extraction, (2) a BEV encoder to transform image features into a BEV representation, and (3) a deformable Transformer decoder that outputs vectorized map instances. 
MapQR enhances efficiency via a GKT-h BEV encoder and scatter-and-gather instance queries for one-to-one decoding in the single-frame regime. MapTracker extends the MapTRv2 decoder with a Strided Memory Fusion module, aggregating temporal features for stable predictions. Despite these differences, the core query-based deformable Transformer decoding structure remains consistent, allowing seamless integration of our proposed priors.


\subsection{Class-aware Spatial Relation Prior}
To better capture structured spatial dependencies between map instances, we introduce a Learnable Class-aware Spatial Relation Prior, which modulates self-attention by incorporating relational cues from geometric and semantic structures. As illustrated in \cref{fig:framework}(b), this prior is computed via a Relation Encoder comprising two components: (1) a Spatial Relation Encoder, which encodes relative positional relationships between instances, and (2) a Class-aware Modulation, which adjusts spatial dependencies based on instance class relationships. As shown in \cref{fig:framework}(a), this prior is integrated into the self-attention mechanism of the Transformer decoder to enhance relation modeling.

\noindent \textbf{Spatial Relation Encoder.} 
Given a set of predicted map instances $\{(x^{l-1}_i, y^{l-1}_i)\}_{i=1}^{N_p}$ from the Transformer decoder layer $l-1$, we compute their relative spatial relationships using bounding box-based positions. To obtain the bounding box for each instance, we use the smallest axis-aligned bounding box that encloses all the $N_p$ points in the instance. Each instance is then represented by its bounding box $\mathbf{b}_i = (cx_i, cy_i, w_i, h_i)$, where $(cx_i, cy_i)$ denotes the center and $(w_i, h_i)$ represents the width and height. This provides a compact representation of the instance’s spatial extent without requiring additional model layers. To encode spatial dependencies, we construct a position relation embedding following \cite{hou2024relation} by computing the relative displacement and scale differences between instances. For each pair of map instances \(i\) and \(j\), we compute their relations:
\begin{align}
\small
\delta x &= \log \left( \frac{|cx_i - cx_j|}{w_i + \epsilon} + 1 \right), \\
\delta y &= \log \left( \frac{|cy_i - cy_j|}{h_i + \epsilon} + 1 \right), \\
\delta w &= \log \left( \frac{w_i}{w_j + \epsilon} \right), \
\delta h = \log \left( \frac{h_i}{h_j + \epsilon} \right),
\end{align}
where $\epsilon$ is a small constant to prevent numerical instability. 

These relative features are then transformed using sinusoidal position encoding and processed through an MLP to obtain a spatial relation embedding:
\begin{equation}
R^{spatial}_{i,j} = f \left( \text{PE} \left( \delta x, \delta y, \delta w, \delta h \right) \right),
\label{eq:positionrelation}
\end{equation}
where $f = \text{ReLU} \circ \text{Linear}$, and $R^{spatial} \in \mathbb{R}^{N_{\text{ins}} \times N_{\text{ins}} \times N_{\text{head}}}$ represents the spatial relationships between all $N_{\text{ins}}$ map instances, with $N_{\text{head}}$ as the number of attention heads. 

\noindent \textbf{Class-aware Modulation.} 
To further refine spatial relationships based on semantic structure, we introduce a \textit{learnable class relation matrix} $\mathbf{R}_{cls} \in \mathbb{R}^{N_c \times N_c \times D}$, which models interactions between different classes. Here, $N_c$ denotes the number of classes, and $D$ is the embedding dimension. This matrix encodes the semantic dependencies between different classes, allowing the model to modulate spatial relationships based on class-specific contextual cues.

Previous works such as \cite{hao2023relation} rely on one-hot assignments to query the class relation matrices, making them highly sensitive to classification errors. Since our model updates relational reasoning based on the classification results from previous layers, inaccuracies in classification predictions in early layers can lead to incorrect relational dependencies, which may propagate through subsequent layers and affect final predictions. To mitigate this, we compute a \textbf{soft class relation prior} using a weighted sum of the learnable class relation matrix, where the predicted class probabilities determine the weights. Specifically, we obtain the predicted class probabilities for each map instance from the previous decoder layer $\mathbf{P}^{l-1}\in \mathbb{R}^{N_{\text{ins}} \times N_c}$, where $P^{l-1}_{i,c_k}$ denotes the probability of instance $i$ belonging to class $c_k$, and $C$ is the total number of classes. Given class probabilities $P^{l-1}_i$ and $P^{l-1}_j$ for map instances $i$ and $j$, the class-aware relation embedding is computed using the following function:
\begin{equation}
R^{cls}_{i,j} = \sum_{c_k=1}^{N_C} \sum_{c_h=1}^{N_C} P^{l-1}_{i,c_k} g(\mathbf{R}_{cls}(c_i, c_j) )P^{l-1}_{j,c_h},
\label{eq:classmod}
\end{equation} 
where $g = \text{ReLU} \circ \text{Linear}$, and $R^{\text{cls}} \in \mathbb{R}^{N_{\text{ins}} \times N_{\text{ins}} \times N_{\text{head}}}$ is class relation embedding, which encodes the pairwise class-aware relational dependencies between instances, capturing how different map elements interact based on their predicted class probabilities.

\noindent \textbf{Integration into Self-attention.} 
The Learnable Class-aware Spatial Relation Prior is obtained by combining the learned spatial relation embedding $R^{\text{spatial}}$ and the class-aware modulation $R^{\text{cls}}$:
\begin{equation}
R^{\text{rel}} = R^{\text{spatial}} + R^{\text{cls}}.
\end{equation}

We incorporate the relation prior into self-attention, where relationships between queries are learned. As shown in \cref{fig:framework} (b), the relation prior is used to modulate the attention weights by adding $R^{\text{rel}}$ to the attention logits before applying the softmax function. The modified attention is formulated as:
\begin{equation}
\text{Attn}_{\text{self}}(\mathbf{Q}, \mathbf{K}, \mathbf{V}) = \text{softmax} \left( \frac{\mathbf{QK}^T}{\sqrt{d}} + R^{\text{rel}} \right) \mathbf{V}.
\end{equation}

By integrating $R^{\text{rel}}$, the self-attention computation is enhanced beyond feature similarities, incorporating structured spatial and semantic dependencies. This improves the model’s ability to capture meaningful relational interactions between instances, leading to more structured and contextually informed predictions. More details of the model structure are provided in the supplementary materials.

\subsection{MoE-based Semantic Prior}

\begin{figure}[t]
    \centering
    \includegraphics[width=\columnwidth]{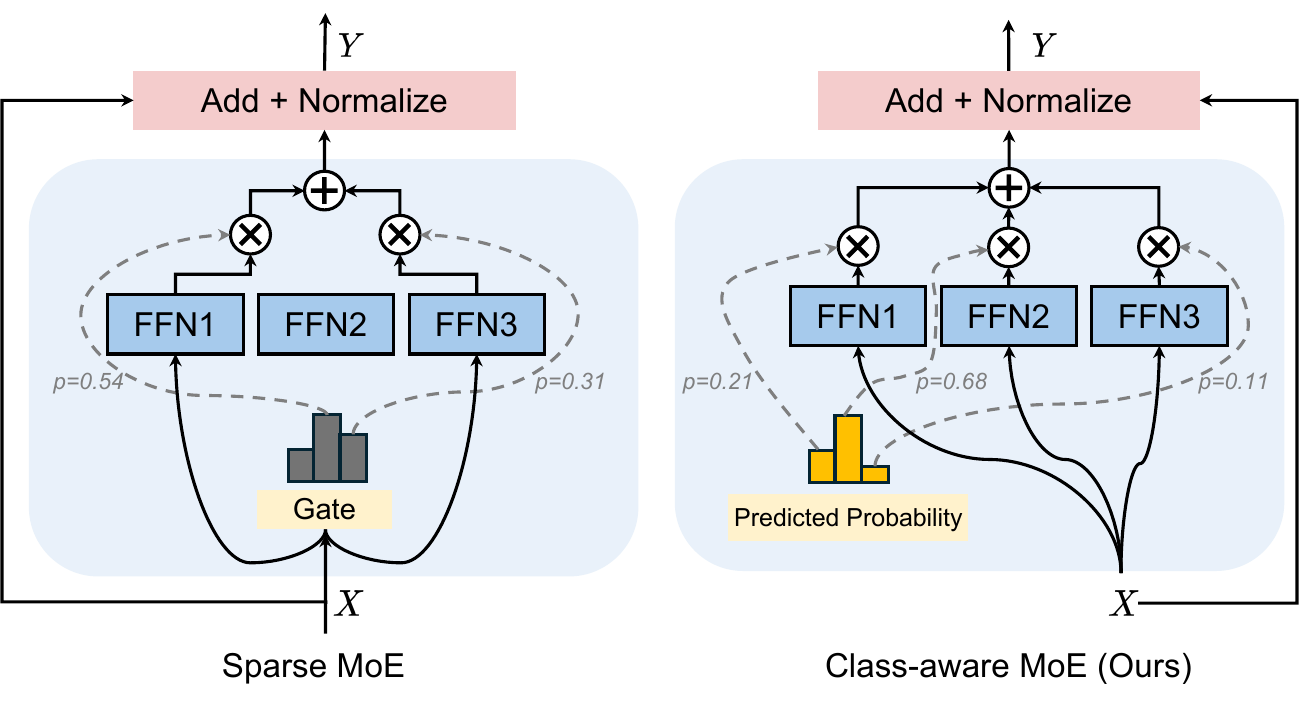}
    \caption{Comparison between \textbf{Sparse MoE} and our \textbf{Class-aware MoE}. In Sparse MoE, a gating network selects the top-k experts and computes a weighted sum of their outputs. In contrast, our method uses predicted class probabilities to assign experts directly, eliminating the need for an additional gating network and allowing each expert to learn class-specific patterns as a semantic prior.}
    \label{fig:moe}
    \vspace{-0.4cm}
\end{figure}

Different types of map elements, such as pedestrian crossings, lane dividers, and road boundaries, exhibit distinct semantic characteristics that influence their contextual roles in HD maps. However, standard Transformer-based models process all instances within a shared feature space, limiting their capacity to capture class-specific nuances. To address this limitation, we introduce a MoE-based Semantic Prior, allowing the model to adaptively refine instance features based on their predicted class probabilities.

As illustrated in \cref{fig:moe}, the MoE module is integrated into the feed-forward network (FFN) of each Transformer decoder layer. Instead of applying a single shared FFN to all instances uniformly, the MoE module dynamically routes instance features to specialized expert networks. The assignment is performed through a weighted combination, where the weights are derived from the predicted class probabilities of the previous decoder layer, $P^{l-1}$, formulated as: 
\begin{equation}
\mathbf{y}_i = \sum_{c=1}^{N_C} P^{l-1}_{i,c} E_c(\mathbf{x}_i),
\end{equation}
where $\mathbf{x}_i$ is the feature of instance $i$, $P^{l-1}_{i,c}$ represents the probability of instance $i$ belonging to class $c$, and $E_c(\cdot)$ is the expert network corresponding to class $c$.

Unlike traditional MoE, which relies on a separate routing network for expert selection, our approach directly utilizes the classification branch output from the previous decoder layer. Traditional MoE architectures require an additional trainable routing network, introducing extra parameters and increasing model complexity. Furthermore, these methods often require an auxiliary loss to balance expert utilization and prevent mode collapse. In contrast, our design eliminates the need for an explicit routing mechanism, reducing computational overhead while ensuring a more stable and interpretable expert selection process. By leveraging the instance classification predictions, our method naturally aligns expert assignment with semantic class distributions, leading to more efficient and semantically coherent feature refinement.

\subsection{Model Training}
\noindent \textbf{RelMap-SF.} Following MapQR~\cite{liu2024mapqr}, we supervise training with three loss terms: a one-to-one instance prediction loss $\mathcal{L}_{\text{one2one}}$, an auxiliary one-to-many loss $\mathcal{L}_{\text{one2many}}$, and a dense foreground segmentation loss $\mathcal{L}_{\text{dense}}$. The overall loss function is defined as:
\begin{equation}
\mathcal{L}_{\text{SF}} = \lambda_o \mathcal{L}_{\text{one2one}} + \lambda_m \mathcal{L}_{\text{one2many}} + \lambda_d \mathcal{L}_{\text{dense}},
\end{equation}
where $\lambda_o$, $\lambda_m$, and $\lambda_d$ are loss weights.

Specifically, $\mathcal{L}_{\text{one2one}}$ employs the Hungarian algorithm with Manhattan distance for bipartite matching, incorporating classification, point-wise, and edge direction losses. $\mathcal{L}_{\text{one2many}}$ introduces an additional prediction branch to generate multiple hypotheses per instance, which improves convergence and robustness. $\mathcal{L}_{\text{dense}}$ includes BEV and PV segmentation losses that guide the model to predict foreground regions across multiple views.

\noindent \textbf{RelMap-TF.} For temporal modeling, we follow MapTracker~\cite{chen2024maptracker} and adopt three objectives: BEV segmentation loss $\mathcal{L}_{\text{BEV}}$, vector tracking loss $\mathcal{L}_{\text{track}}$, and transformation consistency loss $\mathcal{L}_{\text{trans}}$. The total loss is:
\begin{equation}
\mathcal{L}_{\text{TF}} = \mathcal{L}_{\text{BEV}} + \mathcal{L}_{\text{track}} + \lambda_s \mathcal{L}_{\text{trans}},
\end{equation}
where $\lambda_s$ is a balance weight.

Here, $\mathcal{L}_{\text{BEV}}$ combines pixel-wise Focal and Dice losses over rasterized BEV masks. $\mathcal{L}_{\text{track}}$ enforces temporal consistency via hierarchical matching of classification and geometric alignment terms between consecutive frames. Finally, $\mathcal{L}_{\text{trans}}$ ensures that query updates in latent space preserve instance geometry and class semantics over time.
Further architectural and training details are included in the supplementary material.

\section{Experiments}

\begin{table*}[t]
\centering
\caption{Comparison with state-of-the-art methods on nuScenes validation set at 60~m~×~30~m perception range. The APs are calculated with \{0.5m, 1.0m, 1.5m\} thresholds. RelMap-SF achieves the highest mAP among single-frame methods under both 24 and 110 epoch settings, while RelMap-TF outperforms existing temporal frameworks. The evaluation results of other methods are taken from their original papers, except for one marked with \textsuperscript{+}, which is reproduced by MapTracker \cite{chen2024maptracker}.}
\label{tab:nuScenes results}
\footnotesize
\renewcommand{\arraystretch}{1.15}
\setlength{\tabcolsep}{10pt}
\vspace{-0.2cm}
\begin{tabular}{c|c|c|ccc|c|c}
\toprule[1.1pt]
Method & Backbone &Epoch & $\text{AP}_{ped}$ & $\text{AP}_{div}$ & $\text{AP}_{bou}$ & mAP &FPS \\ 
\midrule
HDMapNet \cite{li2022hdmapnet} & EB0 & 30 & 14.4 & 21.7 & 33.0 & 23.0 & -\\
MapTRv2 \cite{liao2024maptrv2} & R50 & 24 & 59.8 & 62.4 & 62.4 & 61.5 & 23.6\\
MGMap \cite{liu2024mgmap} & R50 & 24 & 61.8 & 65.0 & 67.5 & 64.8 & -\\
MapQR \cite{liu2024mapqr}& R50 & 24 & 63.4 & 68 & 67.7 & 66.4 & 22.2\\
HIMap \cite{zhou2024himap}& R50 & 30 & 62.6 & 68.4 & 69.1 & 66.7 & -\\
PriorMapNet \cite{wang2024priormapnet} & R50 & 24 & 64.0 & 69.0 & 68.2 & 67.1 & -\\
MGMapNet \cite{yang2024mgmapnet} & R50 & 24 & 64.7 & 66.1 & \textbf{69.4} & 66.8 & -\\
\cellcolor{gray!20} RelMap-SF (Ours) & \cellcolor{gray!20} R50 & \cellcolor{gray!20} 24 & \cellcolor{gray!20} \textbf{66.3} & \cellcolor{gray!20} \textbf{70.5} & \cellcolor{gray!20} 68.4 & \cellcolor{gray!20} \textbf{68.4} & \cellcolor{gray!20}20.8\\
\midrule
MapTRv2 \cite{liao2024maptrv2} & R50 & 110 & 68.1 & 68.3 & 69.7 & 68.7 & 23.6\\
MGMap \cite{liu2024mgmap} & R50 & 110 & 64.4 & 67.6 & 67.7 & 66.5 & -\\
MapQR \cite{liu2024mapqr} & R50 & 110 & 70.1 & 74.4 & 73.2 & 72.6 & 22.2\\
HIMap \cite{zhou2024himap} & R50 & 110 & 71.3 & 75.0 & 74.7 & 73.7 & -\\
PriorMapNet \cite{wang2024priormapnet} & R50 & 110 & 71.5 & 73.2 & 73.3 & 72.7 & -\\
MGMapNet \cite{yang2024mgmapnet} & R50 & 110 & \textbf{74.3} & 71.8 & \textbf{74.8} & 73.6 & -\\
\cellcolor{gray!20}RelMap-SF (Ours) & \cellcolor{gray!20}R50 & \cellcolor{gray!20}110 & \cellcolor{gray!20}72.2 & \cellcolor{gray!20}\textbf{75.1} & \cellcolor{gray!20}74.7 & \cellcolor{gray!20}\textbf{74.0} & \cellcolor{gray!20}20.8\\
\midrule
StreamMapNet\textsuperscript{+} \cite{yuan2024streammapnet}& R50 &110 & 70.0 & 72.9 & 68.3 & 70.4 & 21.7 \\
MapUnveiler \cite{kim2024unveilinghiddenonlinevectorized} & R50 & 110 & 71.0 & 69.1 & 71.8 & 70.6 & -\\
PrevPredMap \cite{peng2025prevpredmap} & R50 &110 & 71.2 & 70.0 & 72.8 & 71.3 & -\\
MapTracker \cite{chen2024maptracker}& R50 & 72 & 80.0  & 74.1   & 74.1 & 76.1  & 17.6\\
MapExpert \cite{zhang2024mapexpert} & R50 &100   & 79.4 & 73.9   & 76.2 & 76.5 & -\\
HisTrackMap \cite{yang2025histrackmap}& R50  &72 & 79.8 & \textbf{74.5} & 75.4 & 76.6 & -\\
\cellcolor{gray!20} RelMap-TF (Ours)  & \cellcolor{gray!20} R50 &\cellcolor{gray!20} 72 & \cellcolor{gray!20} \textbf{81.1} & \cellcolor{gray!20} 73.6& \cellcolor{gray!20} \textbf{76.6} & \cellcolor{gray!20} \textbf{77.1} & \cellcolor{gray!20} 16.4\\
\bottomrule[1.1pt]
\end{tabular}
\vspace{-0.2cm}
\end{table*}

\subsection{Datasets and Evaluation}
\textbf{Datasets.}
We evaluate our approach on two widely used datasets for HD vectorized map construction: the nuScenes dataset \cite{caesar2020nuscenes} and the Argoverse 2 dataset \cite{wilson2023argoverse}. The nuScenes dataset provides a 360-degree field-of-view (FOV) coverage of the ego-vehicle using six surrounding cameras, capturing diverse urban driving scenarios with 2D vectorized annotations for critical map elements. Argoverse 2 comprises data from seven cameras and provides high-fidelity 3D vectorized map annotations.

\noindent \textbf{Evaluation Metrics.}
Following previous works, we evaluate our method on three types of map instances: lane dividers, pedestrian crossings, and road boundaries. To establish correspondences between predicted and ground truth map elements, we use Chamfer Distance (CD) as the matching criterion under three thresholds: 0.5, 1.0, and 1.5 meters. We compute Average Precision (AP) at each threshold, and report the final mean Average Precision (mAP) as the average over all classes and distance thresholds.

\subsection{Implementation Details}

\noindent\textbf{RelMap-SF.}
We follow the implementation of MapQR~\cite{liu2024mapqr}, using $N_{\text{ins}} = 100$ instance queries, each predicting $N_p = 20$ points. The Transformer decoder is configured with $N_{\text{head}} = 8$ attention heads and an embedding dimension of $D = 256$. For training, we use the AdamW optimizer with a cosine annealing learning rate schedule starting at $6 \times 10^{-4}$. The model is trained for 24 and 110 epochs on nuScenes and 6 epochs on Argoverse 2.
 
\noindent\textbf{RelMap-TF.}
We adopt the training setup of MapTracker~\cite{chen2024maptracker}. During training, four historical frames are randomly sampled from the ten preceding frames (5 seconds). The model is trained for 72 epochs on nuScenes using a three-stage training pipeline and the AdamW optimizer with an initial learning rate of $5 \times 10^{-4}$.

Both models are trained on four NVIDIA L40S GPUs. Additional hyperparameter settings and ablation details are included in the supplementary material.

\begin{figure*}[t]  
    \centering
    \includegraphics[width=\textwidth]{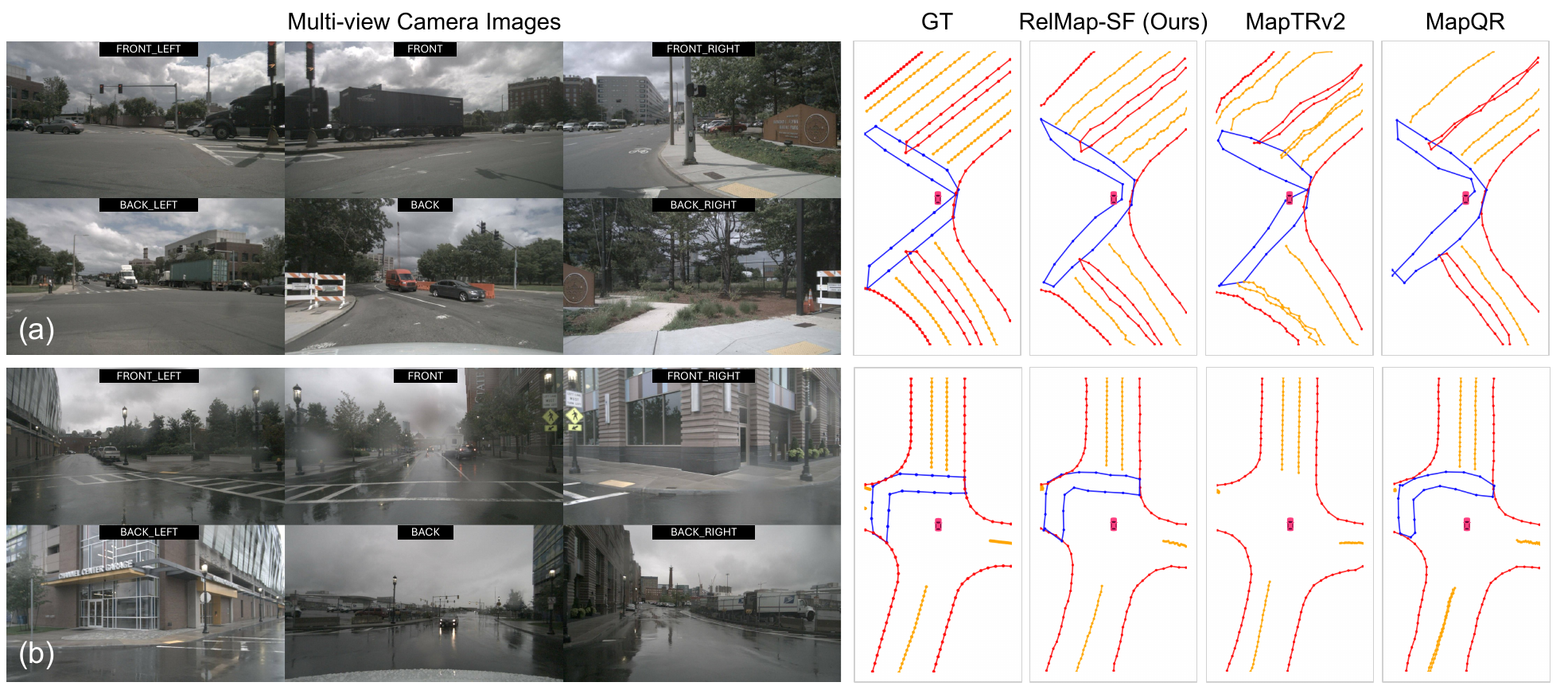} 
    \caption{Qualitative results of RelMap-SF on the nuScenes dataset. The \textcolor{red}{red elements} represent road boundaries, \textcolor{darkyellow}{yellow elements} represent lane dividers, and \textcolor{blue}{blue elements} represent pedestrian crossings. Our model consistently outperforms other models in terms of spatial relations and shape prediction of map elements.}
    \label{fig:qualitative}
    \vspace{-0.4cm}
\end{figure*}

\begin{figure*}[t]  
    \centering
        \includegraphics[width=\textwidth]{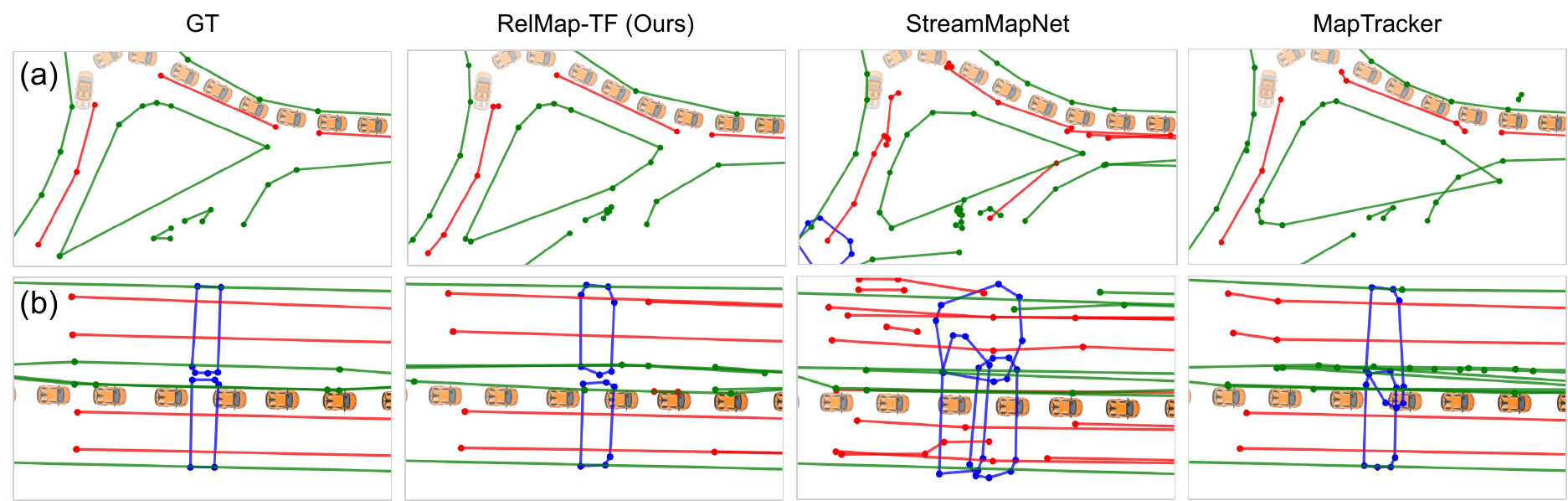} 
    \caption{Qualitative results of RelMap-TF on the nuScenes dataset. The \textcolor{darkgreen}{green elements} represent road boundaries, \textcolor{red}{red elements} represent lane dividers, and \textcolor{blue}{blue elements} represent pedestrian crossings. Our model produces more accurate and spatially consistent map elements.}
    \label{fig:qualitative_tf}
    \vspace{-0.4cm}
\end{figure*}

\subsection{Main Results}


\textbf{Quantitative Results on nuScenes.} 
We evaluate both single-frame and temporal variants of our RelMap framework on the nuScenes \cite{caesar2020nuscenes} validation set. As shown in \cref{tab:nuScenes results}, RelMap-SF consistently achieves the highest mAP among single-frame methods. At 24 epochs, it reaches an mAP of 68.4, outperforming MapTRv2 by 6.9 points. At 110 epochs, it achieves an mAP of 74.0, exceeding MapQR \cite{liu2024mapqr} by 1.4 points and outperforming MapTRv2 by 5.3 points. These substantial gains underscore the efficacy of incorporating spatial and semantic priors.
Compared to PriorMapNet \cite{wang2024priormapnet}, which relies on offline priors that require dataset-specific recomputation, our learnable priors within the Transformer model adapt dynamically during training, eliminating the need for precomputed priors while still achieving a 1.3-point higher mAP than PriorMapNet. 
For temporal map construction, RelMap-TF achieves a new state-of-the-art performance with an mAP of 77.1, outperforming leading methods including MapTracker \cite{chen2024maptracker}, HisTrackMap \cite{yang2025histrackmap}, and MapExpert \cite{zhang2024mapexpert}.
To better assess generalization capabilities, we further evaluate RelMap-TF on the geographically disjoint train/validation split proposed by StreamMapNet \cite{yuan2024streammapnet}, which mitigates the risk of inflated scores due to memorization in the original split. As shown in \cref{tab:newslitnusc}, RelMap-TF achieves an mAP of 42.3, demonstrating the model’s robustness and improved generalization to unseen environments.

These results underscore the effectiveness of our proposed spatial relations and semantic priors in enhancing map construction. The consistent improvements across both single-frame and temporal variants demonstrate that our priors are generalizable and can be integrated into Transformer-based decoders. Furthermore, in terms of computational efficiency, RelMap achieves comparable frames per second (FPS) compared to existing methods, ensuring a balance between accuracy and inference speed.

\begin{table}[t]
  \caption{Comparison with state-of-the-art methods on the geographically disjoint train and validation split of nuScenes proposed by StreamMapNet \cite{yuan2024streammapnet}. }
  \label{tab:newslitnusc}
  \centering
  \vspace{-0.2cm}
  \footnotesize
  \renewcommand{\arraystretch}{1.15}
  \setlength{\tabcolsep}{3.7pt}
  {\begin{tabular}{c|c|ccc|c}
    \toprule[1.1pt]
    Methods & Epoch&  $\text{AP}_{ped}$ & $\text{AP}_{div}$ & $\text{AP}_{bou}$ & mAP \\
    \midrule 
    StreamMapNet \cite{yuan2024streammapnet}& 110 & 31.6 & 28.1 & 40.7 & 33.5  \\
    MapTracker \cite{chen2024maptracker}& 72  &  45.9 & 30.0 & 45.1 & 40.3 \\
    MapExpert \cite{zhang2024mapexpert} & 100 &   46.7 & 34.1 &  45.1 & 42.0 \\
    \cellcolor{gray!20} RelMap-TF (Ours)  & \cellcolor{gray!20}72 & \cellcolor{gray!20} 45.1 & \cellcolor{gray!20}  31.1& \cellcolor{gray!20} 46.7& \cellcolor{gray!20} 42.3\\
  \bottomrule[1.1pt]
\end{tabular}}
\vspace{-0.3cm}
\end{table}

\begin{table}[t]
  \caption{Comparison with state-of-the-art methods on Argoverse 2 validation set. The APs are calculated with \{0.5m, 1.0m, 1.5m\} thresholds. The evaluation results of baseline methods are obtained from the original papers.}
  \label{tab:av2result}
  \centering
  \vspace{-0.2cm}
  \footnotesize
  \renewcommand{\arraystretch}{1.15}
  \setlength{\tabcolsep}{3.7pt}
  {\begin{tabular}{c|c|ccc|c}
    \toprule[1.1pt]
    Methods & Dim  & $\text{AP}_{ped}$ & $\text{AP}_{div}$ & $\text{AP}_{bou}$ & mAP \\
    \midrule 
    VectorMapNet \cite{liu2023vectormapnet} & 2 & 38.3 & 36.1 & 39.2 & 37.9 \\
    MapTRv2 \cite{liao2024maptrv2} & 2  & 62.9 & 72.1 & 67.1 & 67.4 \\
    MapQR \cite{liu2024mapqr}& 2  & 64.3 & 72.3 & 68.1 & 68.2 \\
    HIMap \cite{zhou2024himap}& 2  &\textbf{ 69.0} & 69.5 & \textbf{70.3} & 69.6 \\
    \cellcolor{gray!20} RelMap-SF (Ours) & \cellcolor{gray!20}2 & \cellcolor{gray!20}65.3 & \cellcolor{gray!20} \textbf{74.2} & \cellcolor{gray!20}70.1 & \cellcolor{gray!20} \textbf{69.9} \\
    \midrule 
    VectorMapNet \cite{liu2023vectormapnet} & 3 & 36.5 & 35.0 & 36.2 & 35.8 \\
    MapTRv2 \cite{liao2024maptrv2} & 3  & 60.7 & 68.9 & 64.5 & 64.7 \\
    MapQR \cite{liu2024mapqr} & 3 & 60.1 & 71.2 & 66.2 & 65.9 \\
    HIMap \cite{zhou2024himap}& 3  & \textbf{66.7} & 68.3 & \textbf{70.3} & 68.4 \\
    \cellcolor{gray!20} RelMap-SF (Ours) & \cellcolor{gray!20}3 & \cellcolor{gray!20}64.4 & \cellcolor{gray!20}\textbf{72.7} & \cellcolor{gray!20}68.3 & \cellcolor{gray!20}\textbf{68.5} \\
  \bottomrule[1.1pt]
\end{tabular}}
\vspace{-0.2cm}
\end{table}

\noindent \textbf{Quantitative Results on Argoverse 2.} 
We further evaluate the RelMap-SF model on the Argoverse 2 dataset, which provides 3D (x, y, z coordinates) vectorized map data, allowing assessment in both 2D and 3D. As shown in \cref{tab:av2result}, RelMap-SF achieves an mAP of 68.4 in 2D and 74.0 in 3D, surpassing MapTRv2 by 2.5 and 3.8 points, respectively. These results demonstrate the strong performance of our approach across different spatial representations and datasets.


\noindent \textbf{Qualitative Results.}
We present qualitative comparisons on the nuScenes dataset in \cref{fig:qualitative}. Compared to MapTRv2 \cite{liao2024maptrv2} and MapQR \cite{liu2024mapqr}, our RelMap-SF model produces more accurate, regular, and spatially coherent predictions. In \cref{fig:qualitative}(a), RelMap-SF generates more regular and precise shapes across all three map element classes and better preserves the spatial arrangement between adjacent instances. In contrast, MapTRv2 and MapQR produce less accurate shapes with overlapping or distorted geometry, and MapQR fails to detect certain elements. These improvements demonstrate the effectiveness of the Spatial Relation Prior in enhancing element positioning and maintaining structured layouts. 
In \cref{fig:qualitative}(b), RelMap-SF continues to provide cleaner and more complete predictions, while MapQR and MapTRv2 exhibit deformations or omissions. The adjacency between pedestrian crossings and road boundaries is better preserved in RelMap-SF, whereas in MapQR, the pedestrian crossing and boundary lines intersect incorrectly. 
As shown in \cref{fig:qualitative_tf}, RelMap-TF yields more accurate shapes and spatial arrangements than MapTracker \cite{chen2024maptracker} and StreamMapNet \cite{yuan2024streammapnet}. These results confirm that the proposed spatial and semantic priors remain beneficial in the temporal setting, enabling more stable and coherent predictions across frames.

\subsection{Ablation Studies}
We conduct ablation studies on the nuScenes dataset to assess the effectiveness of individual components and to validate key design choices. All experiments are based on the RelMap-SF model and trained for 24 epochs.

\noindent \textbf{Effectiveness of Key Components.}
We first investigate the impact of the Class-aware Spatial Relation Prior and the MoE-based Semantic Prior by progressively introducing them into the baseline.
As shown in \cref{tab:ablationcomponent}, incorporating the Spatial Relation Prior improves mAP by 0.6, demonstrating its ability to enhance the model’s capacity to capture spatial dependencies among map elements. Adding Class-aware Modulation further enhances mAP by 0.9, demonstrating the advantage of learning spatial relations in a class-specific manner for improved relational reasoning. Finally, using the MoE-based Semantic Prior yields an additional 1.3-point mAP improvement, highlighting its effectiveness in capturing class-specific feature variations. By assigning map instances to specialized experts, the MoE semantic prior enables more adaptive and expressive feature decoding, leading to better performance.

\noindent \textbf{MoE Design Variants.}
To explore different MoE configurations, we experiment with varying the number of experts and the routing mechanisms.  We implement a \textit{Vanilla MoE} following a standard MoE design, where a gating network selects the top-1 expert among three candidates for each instance. 
We also evaluate an MoE design based on MapExpert \cite{zhang2024mapexpert} (\textit{MapExpert MoE}), which expands the number of experts to eight and computes a weighted combination of the top-2 expert outputs. 
Additionally, we implement a variant of our MoE design that incorporates a shared expert (\textit{MoE w/ Shared Expert}) \cite{dai2024deepseekmoe}.
As shown in \cref{tab:ablationmoe}, our MoE design, which utilizes only three experts and does not require an additional routing network, achieves the highest mAP. While MapExpert MoE has a more flexible expert selection mechanism, its performance is sensitive to the auxiliary expert balance loss, necessitating careful hyperparameter tuning. Our approach eliminates the need for additional loss terms, reducing the complexity of tuning while maintaining strong performance. Adding shared expert results in an accuracy drop, indicating that shared feature decoding may weaken the effectiveness of our class-specific prior.


\begin{table}[t]
  \caption{Effectiveness of Key Components in RelMap-SF. $^{\dagger}$: Reproduced result of MapQR \cite{liu2024mapqr} using the official code.}
  \label{tab:ablationcomponent}
  \vspace{-0.2cm}
  \centering
  \footnotesize
  \renewcommand{\arraystretch}{1.15}
  \setlength{\tabcolsep}{3pt}
  \begin{tabular}{cc|c|c}
    \toprule[1.1pt]
    \multicolumn{2}{c|}{Class-aware Spatial Relation Prior} &  Semantic Prior & mAP \\
    \textit{Relation Encoding}& \textit{Class-aware Mod.} & (MoE-based)& \\
    \midrule
    - & - & - & 65.6$^{\dagger}$ \\
     \checkmark & - & - &  66.2  \\
     \checkmark &  \checkmark & - &  67.1  \\
     - & - & \checkmark &  67.3 \\
     \checkmark &  \checkmark &  \checkmark & 68.4 \\
    \bottomrule[1.1pt]
  \end{tabular}
  \vspace{-0.2cm}
\end{table}



\begin{table}[t]
  \caption{Influence of MoE Design Variants in RelMap-SF}
  \label{tab:ablationmoe}
  \vspace{-0.2cm}
  \centering
  \footnotesize
  \renewcommand{\arraystretch}{1.15}
  \setlength{\tabcolsep}{3.7pt}
  {\begin{tabular}{c|cc|c}
    \toprule[1.1pt]
      & \# of Experts & Separate Routing & mAP \\
    \midrule 
    Vanilla MoE & 3  & \checkmark & 66.3  \\
    MapExpert MoE & 8  & \checkmark & 65.9 \\
    MoE w/ shared expert  & 3 + 1 (shared)  &  & 66.7 \\
    \cellcolor{gray!20}Ours & \cellcolor{gray!20}3  & \cellcolor{gray!20} & \cellcolor{gray!20}\textbf{67.3} \\
  \bottomrule[1.1pt]
\end{tabular}}
\vspace{-0.2cm}
\end{table}

\section{Conclusions}
We introduce \textbf{RelMap}, a framework for online vectorized map construction that incorporates a Class-aware Spatial Relation Prior and a MoE-based Semantic Prior. These learnable priors in Transformer-based decoders enhance spatial coherence and class-specific feature refinement, leading to more accurate and structured map representations. We instantiate our framework in two variants: RelMap-SF for single-frame map construction and RelMap-TF for temporal map construction. RelMap models achieve state-of-the-art performance on the nuScenes and Argoverse 2 datasets, highlighting the effectiveness and generalization of the proposed priors. For future work, we aim to explore the integration of external structured knowledge, such as Standard Definition maps, to further guide relational and semantic priors, especially for improving long-range consistency and temporal generalization.

{
    \small
    \bibliographystyle{ieeenat_fullname}
    \bibliography{main}
}

\clearpage
\maketitlesupplementary

\renewcommand{\thesection}{\Alph{section}}
\setcounter{section}{0}

\setcounter{table}{0}
\renewcommand{\thetable}{S\arabic{table}}

\setcounter{figure}{0}
\renewcommand{\thefigure}{S\arabic{figure}}

\setcounter{equation}{0}
\renewcommand{\theequation}{S\arabic{equation}}

\section{Additional Model Details}
\label{sec:modeldetail}
\subsection{Class-aware Spatial Relation Prior}
\noindent \textbf{Spatial Relation Encoder.}
The sinusoidal position encoding is applied to the position relation $(\delta x, \delta y, \delta w, \delta h)$ to generate continuous embeddings that capture spatial relationships between instances. Given the position relation between pairwise instances $R \in \mathbb{R}^{N_{\text{ins}} \times N_{\text{ins}} \times 4}$, the sinusoidal position encoding is computed as follows:
\begin{equation}
\tilde{\mathbf{R}} = \text{PE}(\mathbf{R}) = \left[ \sin\left(\frac{\mathbf{R} \cdot 2\pi}{\mathbf{d}_t} \right), \cos\left(\frac{\mathbf{R} \cdot 2\pi}{\mathbf{d}_t} \right) \right],
\label{eq:sinusoidal}
\end{equation}
where $\mathbf{d}_t$ is a sequence of increasing wavelengths that controls the frequency of the sinusoidal function. It is defined as:
\begin{equation}
d_t(i) = T^{\frac{2i}{D_\text{PE}}}, \quad i = 0, 1, \dots, D_\text{PE}/2 - 1 ,
\end{equation}
where $T$, the temperature parameter that scales the positional encoding, is set to 10,000, and $D_\text{PE}$, the embedding dimension, is set to 16.
As shown in Eq. (2), $f = \text{ReLU} \circ \text{Linear}$ is applied to $\tilde{\mathbf{R}} \in \mathbb{R}^{N_{\text{ins}} \times N_{\text{ins}} \times 4D_\text{PE}}$, transforming it into $\mathbb{R}^{N_{\text{ins}} \times N_{\text{ins}} \times N_\text{head}}$ to match the number of attention heads, allowing it to be incorporated into the self-attention weight later.

\noindent \textbf{Class-aware Modulation.}
We implement the learnable class relation matrix $\mathbf{R}_{cls} \in \mathbb{R}^{N_c \times N_c \times D}$ with $D = 256$. As shown in Eq. (3), we apply $g = \text{ReLU} \circ \text{Linear}$ to transform it into $\mathbb{R}^{N_c \times N_c \times N_\text{head}}$. 

\subsection{RelMap-SF}
\noindent \textbf{Instance Query and Scattered Point Query.}
To better model relationships between map elements, our RelMap-SF model follows MapQR \cite{liu2024mapqr} and adopts a scatter-and-gather query design. This approach employs \textbf{instance queries} for self-attention to model interactions between different map elements while generating \textbf{scattered point queries} for cross-attention to extract fine-grained details from BEV features.

Each map instance is initially represented by a single instance query, which undergoes self-attention to model instance-level relationships. For cross-attention, the \textbf{scatter operation} replicates each instance query into multiple scattered point queries. To ensure that scattered queries retain distinct spatial information, each of them is assigned a \textbf{positional embedding} computed using a positional encoding function that maps reference point coordinates into a high-dimensional space with sinusoidal embeddings. These scattered queries then interact with BEV features through cross-attention, retrieving localized information that captures the geometric structure of the map element.

After cross-attention, a \textbf{gather operation} is used to concatenate all scattered queries corresponding to the same instance and process them through an MLP to form the instance query of the next decoder layer. This step integrates both the global semantic context and the detailed geometric features extracted by the scattered queries, ensuring a comprehensive representation of each map element.

\textbf{BEV Encoder.}
We adopted GKT-h \cite{liu2024mapqr}, which is a variant of GKT \cite{chen2022gkt}, as the BEV encoder in RelMap-SF. GKT (Geometry-guided Kernel Transformer) is a BEV feature extraction method designed to transform multi-view image features into a structured BEV representation. It leverages geometric priors and kernel-based feature unfolding to improve spatial awareness in BEV space. GKT-h \cite{liu2024mapqr} further improves GKT \cite{chen2022gkt} by adaptively predicting the height of 3D reference points for BEV feature projection. Instead of relying on a fixed height, it predicts adjustments of heights based on BEV queries using a linear projection.

\subsection{RelMap-TF}
\textbf{BEV Query Propagation and Memory Fusion.}
We follow MapTracker~\cite{chen2024maptracker} to build the temporal BEV module in RelMap-TF. At each timestep, the BEV feature map is initialized by warping the previous frame’s BEV memory using the ego-motion transformation. To incorporate long-term context, we maintain a temporal buffer of past BEV features and perform memory fusion using a strided selection strategy, where a fixed number of memory slots are selected based on spatial proximity (e.g., 1m, 5m, 10m, 15m away). These selected memories are aligned to the current frame and fused using a lightweight residual update module to produce the final BEV feature for decoding.

\noindent \textbf{Vector Query Propagation and Memory Fusion.}
We adopt the vector memory design from MapTracker~\cite{chen2024maptracker} to support temporally consistent instance decoding in RelMap-TF. At each frame, a set of vector queries is initialized by combining two sources: propagated vectors from tracked elements in the previous frame, and learnable queries for newly emerging elements. The propagated queries are aligned to the current frame through ego-motion transformation and a temporal MLP. To enhance instance-level consistency, a strided selection of vector memories from previous frames is fused with the current vector memory. Each memory-aligned vector undergoes self-attention and BEV-to-vector cross-attention, with updates applied via feedforward networks. This fusion allows the model to retain consistent tracking of road elements across time and improves geometry prediction in dynamic scenes.

\begin{figure*}[tbp]  
    \centering
    \includegraphics[width=0.97\textwidth]{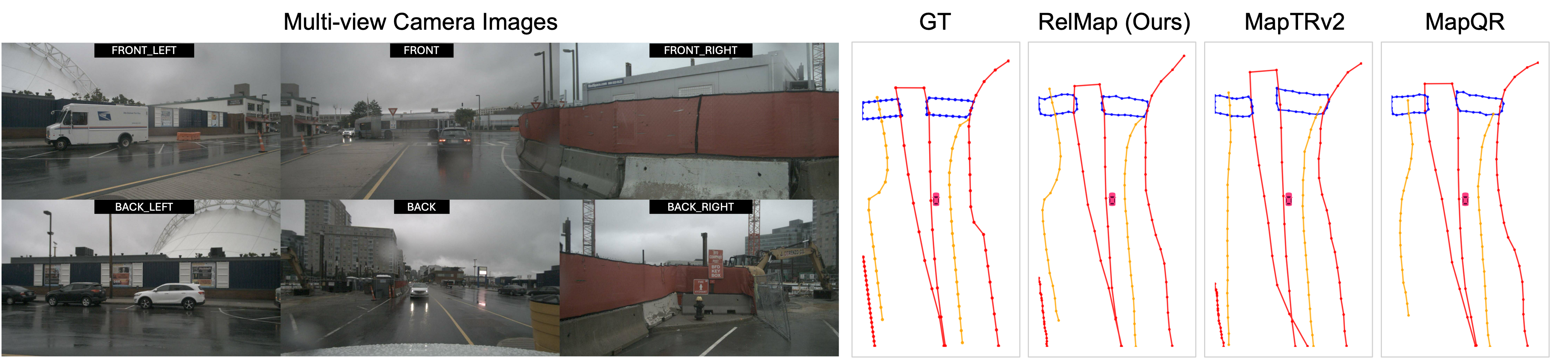}  
    \includegraphics[width=0.97\textwidth]{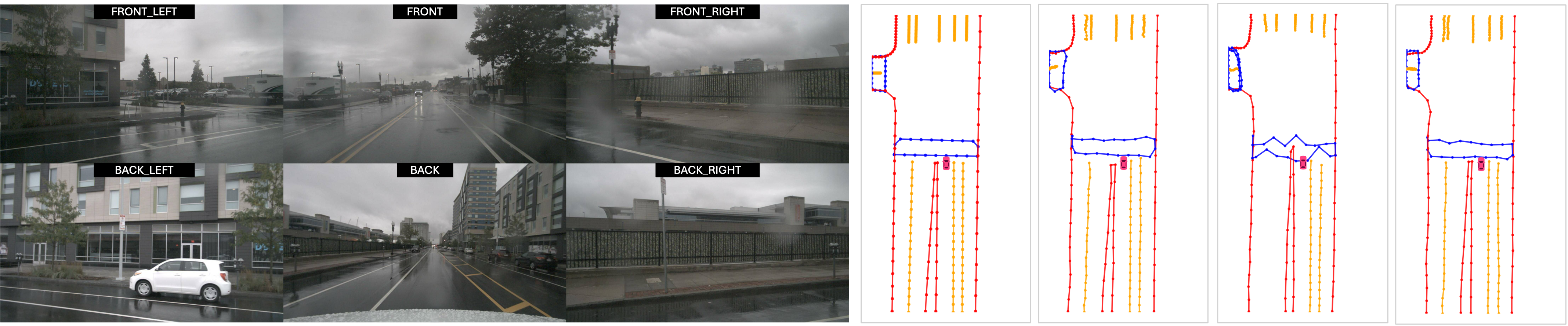}  
    \includegraphics[width=0.97\textwidth]{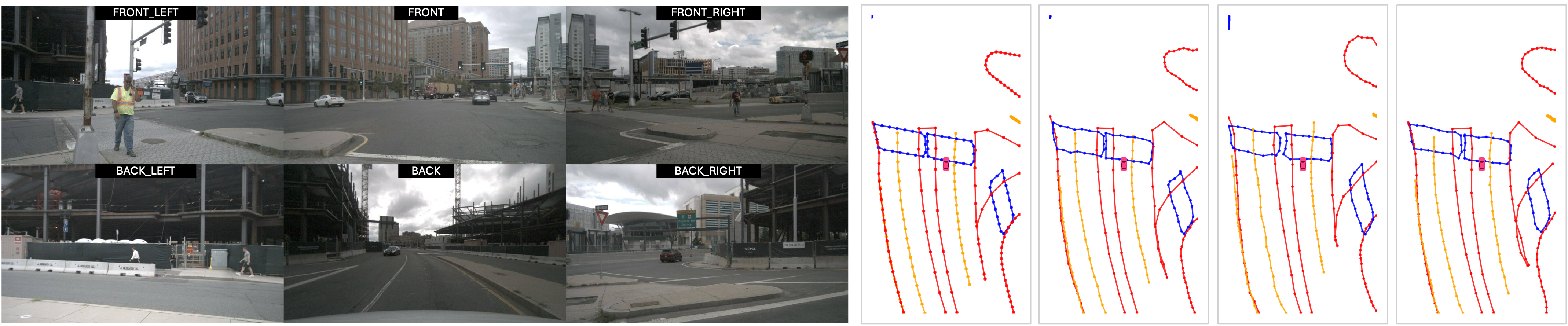}  
    \caption{Additional qualitative results of RelMap-SF on the nuScenes dataset. The \textcolor{red}{red elements} represent road boundaries, the \textcolor{darkyellow}{yellow elements} represent lane dividers, and the \textcolor{blue}{blue elements} represent pedestrian crossings.}
    \label{fig:qualitativesupp}
\end{figure*}

\begin{figure*}[t]  
    \centering
    \includegraphics[width=\textwidth]{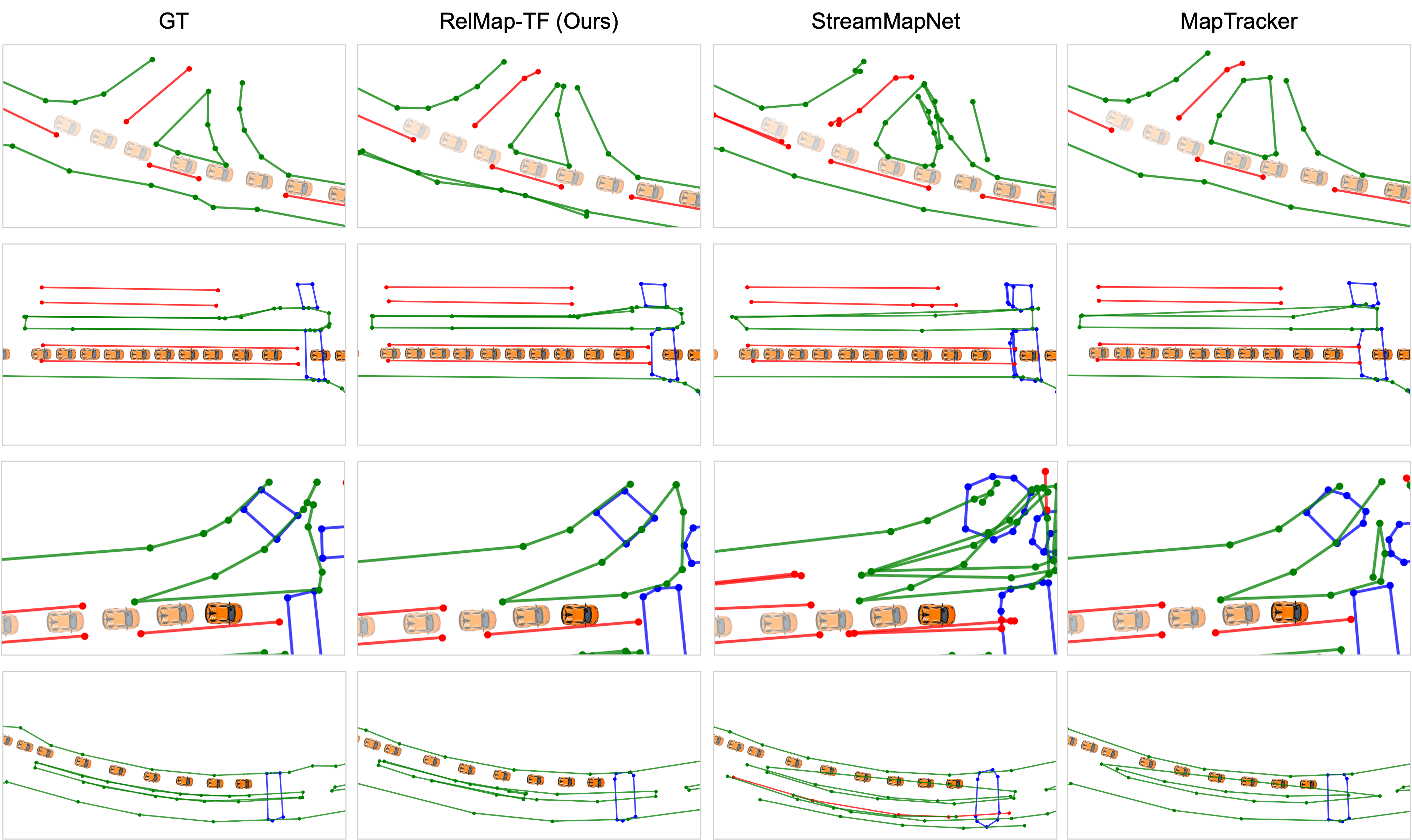} 
    \caption{Additional qualitative results of RelMap-TF on the nuScenes dataset. The \textcolor{darkgreen}{green elements} represent road boundaries, the \textcolor{red}{red elements} represent lane dividers, and the \textcolor{blue}{blue elements} represent pedestrian crossings.}
    \label{fig:qualitativesupp_tf}
\end{figure*}

\section{Model Training}
\subsection{RelMap-SF}
For Relmap-SF, we follow MapTRv2 \cite{liao2024maptrv2} and MapQR \cite{liu2024mapqr} and define training loss as Eq. (7), where $\lambda_o = 1$, $\lambda_m = 1$, and $\lambda_d = 1$ are set in the experiments. 

\noindent\textbf{One-to-one Set Prediction Loss.} 
We employ the Hungarian algorithm \cite{carion2020end} at the instance level to assign predicted instances to ground truth map instances. After instance matching, point-level matching is performed to establish correspondences between predicted and ground truth points using the Manhattan distance. With the matching prediction and ground truth, the $\mathcal{L}_\text{one2one}$ is computed, consisting of a classification loss $\mathcal{L}_\text{cls}$ for map instance category assignment, a point-to-point loss $\mathcal{L}_\text{p2p}$ for aligning predicted and ground truth points' positions, and an edge direction loss $\mathcal{L}_\text{dir}$ to ensure that the predicted edges maintain the same directional order as the ground truth: 
\begin{equation}
\mathcal{L}_\text{one2one} = \alpha_c \mathcal{L}_\text{cls} + \alpha_p \mathcal{L}_\text{p2p} + \alpha_d \mathcal{L}_\text{dir},
\end{equation}
where $\alpha_c = 2$, $\alpha_p = 5$, and $\alpha_d = 0.005$ are set in the experiments. 

\noindent\textbf{Auxiliary One-to-Many Prediction Loss.}
An auxiliary one-to-many matching branch is used during training for faster convergence. This branch shares the same point queries and Transformer decoder as the one-to-one matching branch, but introduces an additional set of instance queries to generate multiple predictions per ground truth instance. Each ground truth instance is duplicated multiple times and padded to form an expanded ground truth set. The Hungarian algorithm is then applied to perform hierarchical bipartite matching between the predicted and expanded ground truth instances. The auxiliary loss $\mathcal{L}_\text{one2many}$ is computed over these matches.

\noindent\textbf{Auxiliary Dense Prediction Loss.} 
The dense prediction loss $\mathcal{L}_\text{dense}$ consists of a BEV segmentation loss $\mathcal{L}_\text{BEVSeg}$, which supervises a predicted foreground mask rasterized onto the BEV canvas, and a PV segmentation loss $\mathcal{L}_\text{PVSeg}$ that supervises a similar mask projected onto perspective-view images. Both the BEV segmentation loss and the PV segmentation loss are implemented as cross-entropy losses. 
\begin{equation}
\mathcal{L}_\text{dense} = \beta_b \mathcal{L}_\text{BEVSeg} + \beta_p \mathcal{L}_\text{PVSeg},
\end{equation}
where $\beta_b = 1$, $\beta_p = 2$ in the experiments.

\subsection{RelMap-TF}

We set the number of queries to 100 and represent each map element using 20 points. We adopt a three-stage training strategy over a total of 72 epochs \cite{chen2024maptracker}. In the first stage (18 epochs), we pretrain the image backbone and BEV encoder. In the second stage (6 epochs), we enable the vector decoder while freezing the previously trained modules. To ensure stable initialization, the vector memory is activated after 500 warmup iterations. In the final stage (48 epochs), we jointly fine-tune the entire model with all modules active and supervised by the full loss $\mathcal{L}_{\text{TF}}$.

\noindent\textbf{Overall Loss.}
To enable temporally consistent prediction, RelMap-TF adopts the training strategy of MapTracker~\cite{chen2024maptracker}, with the overall training loss as:
\begin{equation}
\mathcal{L}_{\text{TF}} = \mathcal{L}_{\text{BEV}} + \mathcal{L}_{\text{track}} + \lambda_s \mathcal{L}_{\text{trans}},
\end{equation}
where $\lambda_s$ is a weighting factor for the transformation loss and is set to 0.1 in the experiments. The three loss components are described below.

\noindent\textbf{BEV Segmentation Loss.}
This term supervises the model to predict accurate BEV foreground masks from multi-view inputs. It combines a focal loss and a dice loss over the predicted binary segmentation map.
\begin{equation}
\mathcal{L}_{\text{BEV}} = \lambda_f \mathcal{L}_{\text{focal}}(\mathbf{S}(t), \hat{\mathbf{S}}(t)) + \lambda_c \mathcal{L}_{\text{dice}}(\mathbf{S}(t), \hat{\mathbf{S}}(t)),
\end{equation}
where $\lambda_f$ is set to 10.0, and $\lambda_c$ is set to 1.0.

\noindent\textbf{Vector Tracking Loss.}
To supervise instance-level predictions across time, we adopt a loss that combines classification and geometry consistency. For new elements, predictions are matched to ground truth via bipartite matching using a composite cost over class labels and vector shapes. For tracked elements, the matching from the previous frame is reused. The total tracking loss is the sum of a focal loss on predicted categories and a permutation-invariant line loss on vector coordinates.

\noindent\textbf{Transformation Consistency Loss.}
This term regularizes propagated vector queries by enforcing that their updated representations remain consistent with expected geometric and semantic changes under ego-motion. It improves temporal coherence during memory fusion and propagation.

\section{Additional Results}
\subsection{Ablation Studies}
\textbf{Instance Position Representation}
We investigate two different methods for encoding the spatial relations of instances:
(1) Bounding Box Representation: A map instance is represented by the smallest axis-aligned bounding box that encloses all its points. Given an instance with points $\{(x_i, y_i)\}{i=1}^{N_p}$, the bounding box is defined by its center and dimensions as follows:
\begin{align}
(cx, cy, w, h) &= \left( \frac{x_{\min} + x_{\max}}{2}, \frac{y_{\min} + y_{\max}}{2}, \right. \notag \\
&\quad \left. x_{\max} - x_{\min}, y_{\max} - y_{\min} \right).
\end{align}

(2) Four-point Representation: A map instance is represented by a set of four key points selected at fixed intervals along its sequence of points. Specifically, the four points $(x_0, y_0)$, $(x_6, y_6)$, $(x_{12}, y_{12})$, and $(x_{19}, y_{19})$ are chosen from the $20$ points that define each instance.

\begin{table}[t]
  \caption{Influence of different instance position representation on Class-aware Spatial Relation Prior in RelMap-SF. The MoE-based Semantic Prior is excluded from this experiment. The models are trained and evaluated on the nuScenes dataset.}
  \label{tab:ablationpoints}
  \centering
  \footnotesize
  \renewcommand{\arraystretch}{1.15}
  \setlength{\tabcolsep}{3.7pt}
  {\begin{tabular}{c|c|c}
    \toprule[1.1pt]
    Position Rep.      & Epoch & mAP \\
    \midrule 
    Four-point Repr.    & 24    & 66.5 \\
    Bounding Box Repr.  & 24    & \textbf{67.1}  \\
  \bottomrule[1.1pt]
\end{tabular}}
\end{table}

To evaluate the impact of the instance position representation, we train RelMap-SF with only the Class-aware Spatial Relation Prior, excluding the MoE-based Semantic Prior. As shown in \cref{tab:ablationpoints}, the Bounding Box Representation yields a higher mAP compared to the Four-point Representation, thus making it the preferred choice in this paper.

\noindent\textbf{Positional Encoding in Spatial Relation Encoder.} 
We compare a non-learnable sinusoidal positional encoding, as described in \cref{sec:modeldetail}, with a learnable Fourier-based embedding for spatial relation encoding. The Fourier-based embedding extends the sinusoidal encoding by introducing learnable frequency components instead of using a fixed set of wavelengths. Rather than directly applying predefined frequencies, it learns frequency embeddings through a parameterized embedding layer. The Fourier-based positional encoding is formulated as follows:
\begin{equation}
\tilde{\mathbf{R}} = \left[ \sin\left( \mathbf{R} \cdot \mathbf{W}_\text{PE} \right), \cos\left( \mathbf{R} \cdot \mathbf{W}_\text{PE} \right), \mathbf{R} \right],
\end{equation}
where $\mathbf{W}_\text{PE} \in \mathbb{R}^{4 \times D_\text{PE}}$ is a learnable frequency matrix that replaces the fixed wavelength sequence $\mathbf{d}_t$ in the sinusoidal encoding, as defined in \cref{eq:sinusoidal}.

To evaluate the impact of learnable versus non-learnable positional encoding, we conduct an ablation study on RelMap-SF using only the Class-aware Spatial Relation Prior. As reported in \cref{tab:ablationpe}, the non-learnable encoding achieves a slightly higher mAP while reducing model complexity. Therefore, we adopt the non-learnable positional encoding in our final model.

\begin{table}[t]
  \caption{Comparison of non-learnable and learnable positional encoding for spatial relation encoding. The MoE-Based Semantic Prior is not incorporated in this experiment. The models are trained and evaluated on the nuScenes dataset.}
  \label{tab:ablationpe}
  \centering
  \footnotesize
  \renewcommand{\arraystretch}{1.15}
  \setlength{\tabcolsep}{3.7pt}
  {\begin{tabular}{c|c|c}
    \toprule[1.1pt]
    PE Method & Epoch & mAP \\
    \midrule 
    Learnable PE   & 24 &  67.0 \\
    Non-learnable PE & 24 & \textbf{67.1} \\
  \bottomrule[1.1pt]
\end{tabular}}
\end{table}

\subsection{Data Efficiency}
\begin{figure}[t]  
    \centering
    \includegraphics[width=\columnwidth]{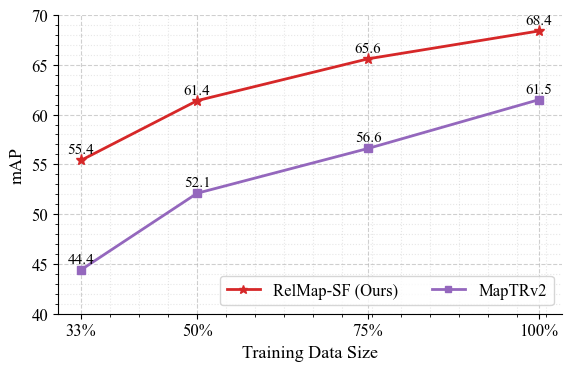}
    \caption{Comparison of model performance (mAP) with varying training data sizes on the nuScenes dataset. RelMap-SF (ours) achieves competitive accuracy with only 50\% of the training data, matching the performance of MapTRv2 trained on the full dataset. }
    \label{fig:dataefficiency}
\end{figure}

We show that using the proposed prior relation learning in RelMap enhances training efficiency, particularly in limited-data scenarios \cite{wang2020fewshot}.  To assess the effectiveness of our proposed priors under limited data, we conduct experiments using 75\%, 50\%, and 33\% of the full nuScenes \cite{caesar2020nuscenes} training dataset. As shown in \cref{fig:dataefficiency}, our model consistently outperforms competing approaches across all data sizes. Notably, with only 50\% of the training data, our method achieves an mAP of 61.4, matching the MapTRv2 model when trained on the entire dataset.

\subsection{Qualitative Results}
\cref{fig:qualitativesupp} presents additional qualitative results in complex scenarios on the nuScenes dataset, comparing our RelMap-SF model with baseline methods MapTRv2 \cite{liao2024maptrv2} and MapQR \cite{liu2024mapqr}. \cref{fig:qualitativesupp_tf} shows qualitative results of RelMap-TF on the nuScenes dataset, highlighting comparisons with baseline temporal online mapping models, including StreamMapNet \cite{yuan2024streammapnet} and MapTracker \cite{chen2024maptracker}. The results demonstrate that our RelMap models consistently generate higher-quality maps, outperforming the baseline methods in both single-frame and temporal settings.

\end{document}